\documentclass[letterpaper, 10 pt, conference]{ieeeconf}  % Comment this line out if you need a4paper

\IEEEoverridecommandlockouts                              % This command is only needed if 
\usepackage{graphics} % for pdf, bitmapped graphics files
\usepackage{xcolor}
\usepackage{hyperref}
\usepackage{epsfig} % for postscript graphics files
\usepackage{mathptmx} % assumes new font selection scheme installed
\usepackage{times} % assumes new font selection scheme installed
\usepackage{amsmath} % assumes amsmath package installed
\usepackage{amssymb}  % assumes amsmath package installed
\usepackage{multirow}
\usepackage{subfig}
\usepackage{lipsum}
\usepackage{afterpage}
\usepackage{caption}
\usepackage{makecell}
\usepackage{multirow}
\usepackage{booktabs}
\DeclareCaptionFont{capsize}{\fontsize{8}{8.8}\selectfont}
\renewcommand{\baselinestretch}{0.95}
\title{\LARGE \bf
Learning Traversability for Long Horizon Off-Road Navigation
}

\author{Kasi Viswanath$^{1}$, Jason M. Gregory$^{2}$, Shaunak Kolhe$^{1}$ and Srikanth Saripalli$^{1}$% <-this % stops a space
\thanks{*This research was developed with funding from the Defense Advanced Research Projects Agency (DARPA) and DEVCOM Army Research Laboratory (ARL) and was accomplished under Cooperative Agreement Number W911NF-21-2-0064. The views, opinions and/or findings expressed are those of the authors and should not be interpreted as representing the official views or policies of DARPA, ARL or the U.S. Government.}
\thanks{$^{1}$J. Mike Walker '66 Department of Mechanical Engineering, Texas A\&M University, College Station, TX 77840, USA
        {\tt\small kasiv, kolheshaunak, ssaripalli@tamu.edu}}%
\thanks{$^{2}$DEVCOM Army Research laboratory, Adelphi, MD, USA
        {\tt\small  jason.m.gregory1.civ@army.mil}}%
}

\begin{document}

\maketitle
\thispagestyle{empty}
\pagestyle{empty}

\begin{abstract}
Autonomous navigation across large off-road environments remains a challenging problem. Onboard sensors perceive only the immediate surroundings, yet safe and efficient routes depend on terrain features that extend well beyond the sensor horizon. Geo-spatial data sources such as satellite imagery, aerial LiDAR, and vector maps can close this gap, but learning traversability from them is difficult: dense labels are unavailable at scale, and existing methods rely on short-range sensing. We propose an efficient formulation that learns a continuous traversability map from overhead data, supervised directly by human-driven GPS trajectories and shaped by supervised geometric priors from LiDAR. Alongside the model, we release a dataset, curated from public sources, consisting of 299 scenes spanning $\sim\!1{,}244\,\mathrm{km}^{2}$ of diverse terrain, paired with $1{,}130\,\mathrm{km}$ of human driving. In field trials on a Clearpath Warthog across seven routes at two sites,our method achieves trajectories within $5.5\%$ of human path length and reduces operator interventions by $\sim\!85\%$ compared to local-planner-only autonomy.~\textcolor{blue}{\url{https://unmannedlab.org/research/long-horizon-traversability}}
\end{abstract}

\section{Introduction}

Navigation in urban environments is inherently hierarchical: global route planners derived from map services produce road-level decisions optimized for travel time, traffic, or energy, which are then continuously refined by reactive local planners that respond to immediate context~\cite{paden2016survey}. While this paradigm has been extensively studied and deployed at scale, off-road environments remains relatively underexplored.

Off-road lacks well-defined road networks, exhibit significant spatial variability, and often contain noisy or incomplete prior information. Feasible routes may not correspond to visible trails, and existing paths may be suboptimal or impassable~\cite{schmittle2025lrn,damm2023terrain,talia2025incremental}. As a result, planning must operate directly over terrain representations, placing a strong emphasis on accurate traversability estimation over large spatial extents. Traversability maps are central to this process, encoding the safety and feasibility of regions while enabling the generation of diverse, goal-directed routes.

% \begin{figure}[t]
%     \centering
%     \includegraphics[width=0.85\linewidth]{images/warthog_exp.png}
%     \caption{Warthog in action across diverse off-road terrain, 
%     demonstrating long-horizon autonomous navigation.} 
%     \label{fig:warthog}
%     \vspace{-0.7cm}
% \end{figure}

Learning a costmap from overhead data is difficult because dense traversability
labels do not exist at scale. Driven trajectories offer a way around this with each
traversed cell is direct evidence that the terrain was passable, requiring no
annotation. Existing work either infers traversability from terrain structure
without reference to where vehicles drive, or learns from driving data but
recovers the cost by backpropagating through a planner, which restricts the
planners that can be used at deployment. This raises a
fundamental question: \emph{can a robot learn from overhead data a costmap that
is grounded in measured terrain geometry and calibrated by where vehicles have
actually driven?}

We answer this  by introducing a global planning framework that jointly learns and exploits traversability-aware representations for off-road navigation. Our unified model integrates heterogeneous geospatial data such as satellite imagery, aerial LiDAR, vector maps and predicts both a continuous traversability representation and a goal-conditioned estimate of route preference. By supervising with driven GPS trajectories via a Positive-Unlabeled formulation~\cite{kiryo2017positive}, the framework treats traversed terrain as verified traversable and unvisited terrain as unknown rather than lethal, while still producing a dense per-pixel map registered to the input grid. This design captures physical feasibility alongside human navigation behavior without requiring differentiable planners or manual annotation.

Our main contributions are:
\begin{itemize}
    \item A publicly sourced multi-modal geospatial dataset spanning $\sim$1{,}244\,km$^2$ across 299 unique scenes with diverse terrain types, paired with 1{,}130\,km of driven GPS trajectories, providing the largest geographic coverage for off-road navigation to our knowledge.

    \item A goal-conditioned path-likelihood learning framework supervised by driven trajectories via a Positive-Unlabeled formulation.
    % , treating traversed terrain as verified traversable without requiring a differentiable planner.

    \item A continuous-valued traversability map learned from shared representations and shaped by weakly weighted LiDAR-derived anchors (e.g., slope, height, intensity), enabling dense terrain understanding.

    \item Extensive real-world validation on a Clearpath Warthog across a levee site and an off-road site, where our Long-Range Planner (LRP) achieves trajectories within 5.5\% of human baseline distance (vs.\ 13.9\% for a Short-Range Planner (SRP) baseline) and reduces operator interventions by 85\%.
\end{itemize}

\section{Related Work}\label{sec:related}

\subsection{Off-road Datasets}

The development of off-road navigation datasets has largely followed advances in learning-based perception. Early datasets such as RUGD~\cite{RUGD2019IROS} focused on semantic understanding from RGB images, while later efforts including RELLIS-3D~\cite{9561251}, GOOSE~\cite{goose-dataset} and GO~\cite{jiang2025gogreatoutdoorsmultimodal} incorporated multi-modal sensor with LiDAR, IMU, GPS, and RADAR to support tasks such as localization, and local navigation.

TartanDrive 2.0~\cite{sivaprakasam2024tartandrive} adds driving trajectories for cost map learning, but like all onboard collection they capture detailed observations only along traversed paths, offering limited spatial coverage for long-range planning.

Geospatial data offers a complementary perspective by providing large-scale coverage of terrain through satellite imagery, aerial LiDAR, and digital elevation models. Public sources including ArcGIS~\cite{esri_world_imagery}, USGS~\cite{usgs_lidar_2024}, and OpenStreetMap~\cite{OpenStreetMap} make such data widely accessible, but existing geospatial datasets are built for semantic understanding rather than navigation and carry no traversability supervision.

\subsection{Terrain Traversability}

Traversability estimation has traditionally been studied for local navigation.
Cost maps were built from hand-engineered features and semantic class-to-cost
association~\cite{Montemerlo2006WinningTD,manduchi2005obstacle}, later from
learned perception and proprioception~\cite{https://doi.org/10.1002/rob.20168},
and more recently from vehicle dynamics and execution feedback~\cite{hdif2023}.
These methods operate within the sensor horizon.

Overhead data extends that range. Early work used aerial imagery and LiDAR with
simple models~\cite{silver2008high}, and hierarchical planners for planetary
exploration derive cost analytically from orbital
priors~\cite{toupet2026perseverance,richter2024lunar}.
OVerSeec~\cite{rana2026overseec} instead assigns cost by segmenting satellite
imagery and classifying the segments with a vision-language model against a
prompt, without training on the task.
Trailblazer~\cite{viswanath2025trailblazerlearningoffroadcostmaps} learns the
cost from driving data, combining satellite semantics with aerial LiDAR through
a differentiable planner~\cite{pmlr-v139-yonetani21a}. Recent work pushes that horizon outward, predicting multi-resolution traversability and elevation to ±100m from onboard cameras and LiDAR, with satellite DEMs used to complete the elevation labels rather than as a network input~\cite{patel2024roadrunner}. The horizon is extended but remains egocentric and bounded by the vehicle's own sensing.

Learning cost from driving data places a planner in the training loop. Inverse
and deep reinforcement learning
approaches~\cite{wulfmeier2017large,9812238,zhu2020offroad} solve the planning
problem at every iteration, which presumes a tractable state space and a model of
the demonstrating agent. Approaches that backpropagate through the planner require it to be differentiable, excluding search methods with
completeness guarantees~\cite{Hart1968,yen1971finding} and the kinodynamic
planners used off-road~\cite{damm2023terrain,talia2025incremental}.

What remains is a costmap grounded in measured terrain appearance and geometry and calibrated against where vehicles have driven. That
combination, learned at scale from overhead data without a planner in the
training loop, is the gap this work addresses.
\begin{figure*}[ht]
    \centering
    \includegraphics[width = 0.9\textwidth, height = 5 cm]{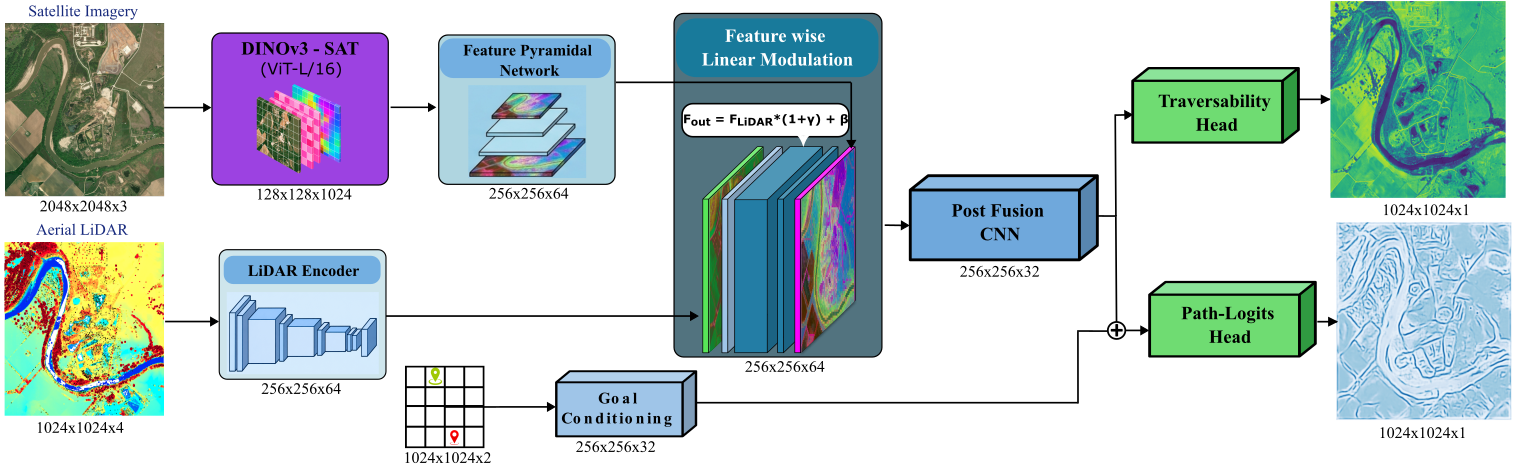}
    \caption{Architecture: Satellite imagery is encoded by a frozen DINOv3-SAT ViT-L/16 and decoded by an FPN. LiDAR-
derived height, slope and intensity rasters and an OSM channel are encoded by a CNN. The two streams are
fused by spatially-varying FiLM. The goal heatmaps are encoded separately and reach the path-logits head only,
so the traversability head is goal-agnostic. Tensor shapes are given below each block.} 
    %Traversability maps and cost maps are visualised using the {\color[HTML]{35B779}\emph{viridis}} and  {\color[HTML]{B12A90}\emph{inferno}} colormaps respectively.}
    \label{fig:architecture}
    \vspace{-0.5cm}
\end{figure*}

\section{Data Collection and Processing}\label{data_collection}
Effective off-road navigation requires reasoning over heterogeneous terrain cues that no single sensor can fully capture. High-resolution satellite imagery provides rich visual context, vegetation, trails, and water bodies, yet cannot resolve elevation or slope, which aerial LiDAR recovers directly. OSM's vector layer complements both by supplying map annotations of highways, trails, and waterways as strong semantic priors for the scene. OpenStreetMap's also hosts a public GPS trace corpus, a separate crowd-sourced repository of raw, user-uploaded trajectories. We use these traces as evidence of traversability through the terrain.

We identified candidate off-road sites spanning a broad range of terrain types such as deserts, grasslands, evergreen forests, mines, and mountainous regions, cross-referenced against the availability of GPS traces in this corpus. After a feasibility study, we retained only regions where all three modalities were jointly available alongside sufficient trace coverage. Each candidate trace is verified against the satellite imagery and filtered by speed derived from consecutive-fix timestamps, discarding stationary segments and non-vehicular activity such as walking. Of the retained corpus, $67.5\%$ of traversed distance falls outside a $5\,\mathrm{m}$ buffer of any OSM road or trail centreline.

The U.S. Geological Survey (USGS)~\cite{usgs_lidar_2024} distributes overhead LiDAR point clouds, covering more than 85\% of the U.S. land area with a vertical accuracy of 10\,cm (4\,in). These point clouds are dense, with 5--27 points per $m^{2}$, and include semantic classification in many regions. Data from the USGS LiDAR Explorer~\cite{usgs_lidar_2024} is organized into \textit{tiles}, \textit{swaths}, or \textit{projects}, typically at 1\,km or 1.5\,km grid resolution.

For each selected region we downloaded the point cloud, then used its geographic bounds to extract co-registered 30\,cm/pixel satellite imagery from ArcGIS~\cite{esri_world_imagery}, along with OSM vector features and the filtered GPS traces described above. Data acquisition is spread across seasons to capture natural variation in vegetation and surface conditions.

The resulting dataset comprises \textbf{299 unique scenes} spanning approximately \textbf{1{,}244\,km\textsuperscript{2}}, paired with \textbf{1{,}130\,km} of driven GPS traces. Together, these modalities form the input to our algorithm's preprocessing pipeline, described next.

\section{Methodology}
Having established the data sources in Section~\ref{data_collection}, we now describe how these heterogeneous geospatial inputs are transformed into a learned traversability representation. We first detail the preprocessing pipeline that aligns and rasterizes the inputs into grid-maps, then present the architecture that fuses them into path-likelihood and traversability map predictions.

\subsection{Data pre-processing}\label{sec:preprocessing}

LiDAR point clouds contain tens of millions of points, so we rasterize each into
aligned height, slope, and intensity channels. Rasterization discards fine 3-D
structure but yields a fixed-size representation at the scale of the satellite
imagery. The grid resolution $G_{\text{res}}$ is set from local point density to
ensure at least $15$ points per cell, the minimum for stable surface-normal
estimation, and the cloud is voxelized at $0.25 \cdot G_{\text{res}}$ to
regularize point distribution. Slopes are computed as the angle
$s = \arccos{(n_z)}$, where $n_z$ is the vertical component of the surface
normal.

LiDAR intensity acts as a helpful visual guide. Vegetation and water absorb strongly and return low intensity, while trails, roads, and barren ground return high. To align this channel with the ``higher value $\Rightarrow$ less traversable'' convention used for height and slope, we normalize raw intensity $I \in[0,1]$ and store its complement $\tilde{I} = 1 - I$ on the grid map.

The three channels are rasterized at $G_{\text{res}}$, with each cell storing the mean of all enclosed points. OpenStreetMap features (highways, trails, waterways) are rasterized onto the same grid to provide semantic priors. Corresponding $30\,\mathrm{cm/pixel}$ satellite imagery is cropped and reprojected to match the LiDAR extents under a common datum (NAD83 UTM), preserving spatial consistency and precise cross-modal alignment.

GPS traces from each scene are augmented to generate trajectories ranging from $200\,\text{m}$ to $1.5\,\text{km}$ in length. Each trajectory is rasterized to $G_{\text{res}}$, with occupied cells set to $1$ and background cells to $0$. Start and goal points are subsequently extracted to construct a separate start--goal map. 
\subsection{ Model Architecture}

An overview of the proposed architecture is shown in Fig.~\ref{fig:architecture}, which is a dual-stream, multi-modal network that jointly processes satellite imagery and LiDAR-derived grid maps to produce a traversability map and path logits.

\subsubsection{Satellite Image Encoder \& Feature Pyramidal Network(FPN)}
Satellite imagery is encoded by DINOv3-SAT, a ViT-L/16 pretrained on SAT-493M~\cite{simeoni2025dinov3}. 
%Final-layer patch tokens are reshaped into a $64{\times}64$ grid ($p{=}16$, $d{=}1024$). 
The encoder is kept frozen, as its satellite-domain pretraining provides semantically rich land-cover, vegetation, and geometry features that are difficult to recover from task-specific supervision alone. A FPN decodes the DINO features to fusion resolution. The finest level is progressively upsampled and compressed to $64$ channels, preserving both fine spatial detail and abstract semantic context at fusion.

\subsubsection{LiDAR Encoder}
The rasterized LiDAR grid maps of height, slope, intensity, and an OSM raster channel are encoded by a three-stage Convolution Neural Network (CNN) yielding $64$-channel features at full resolution. Learnable per-channel input scales are applied prior to encoding with the features then average-pooled to fusion resolution.

\subsubsection{Multi-Modal Fusion via FiLM}
LiDAR and satellite features are fused via spatially-varying Feature-wise Linear Modulation (FiLM)~\cite{10.5555/3504035.3504518}. The satellite-derived DINO features predict a per-cell modulation map, $\gamma, \beta \in \mathbb{R}^{C \times H \times W}$, rather than the single global pair used in the original formulation, and this map reweights LiDAR features at every spatial location:
\begin{equation}
    \mathbf{F}_{\text{fused}} = \mathbf{F}_{\text{LiDAR}} \odot (1 + \gamma) + \beta
\end{equation}
Intuitively, satellite context at each cell governs how strongly the corresponding LiDAR feature is amplified, suppressed, or shifted. Under dense canopy, where ground returns are less reliable, LiDAR intensity is attenuated whereas in open terrain it passes through largely unchanged. $\gamma \in (-1,1)$ is $\tanh$-bounded to keep this reweighting stable, and a post-fusion block further refines the output.

\subsubsection{Goal Conditioning}
Start and goal positions are provided as spatial heatmaps, encoded by a two-layer CNN into $32$-channel goal features, which are concatenated with the $64$-channel fused representation for the path-logits head. The encoding is spatial rather than a vector embedding so that it remains aligned with the dense decoder output. At inference the start/goal channel may be left empty, as it is in all field experiments reported in Sec.~VII.

\subsubsection{Post-Fusion Refinement and Output Heads}
The fused representation is refined by a shallow CNN and feeds two heads: a
\textbf{Traversability Head} taking the $64$-channel refined features, and a \textbf{Path Logits Head} taking those features concatenated with the
$32$-channel goal encoding. The goal features are therefore routed to the path head only, and the traversability head is goal-agnostic . Each head comprises a two-stage conv block with a single output channel.

The two outputs answer different questions. The \emph{traversability map}
$\hat{T} \in [0,1]$ is a per-pixel estimate of how readily a cell can be
traversed, independent of any particular journey. It is sigmoid-normalised and is the representation consumed by the planner. The \emph{path logits} are unnormalised per-pixel scores whose sigmoid $\hat{p}$ gives the likelihood that a cell lies within a corridor traversed between the given start and goal. They are a goal-conditioned route prediction rather than a terrain property. Both outputs are bilinearly upsampled to $1024{\times}1024$ prior to loss computation.
\subsection{Loss Function}
The network is trained with a composite loss. Path supervision acts on the
path-logits head, while the traversability head is shaped by weak positive
signals and supervised geometric priors, described below. Direct traversability supervision is infeasible as dense per-pixel labels are unavailable, and unvisited pixels cannot be assumed non-traversable. 

\subsubsection{Path Corridor Loss}

Path supervision follows a Positive--Unlabeled (PU)
formulation~\cite{kiryo2017positive} where pixels within a band of radius
$r=5\,\text{px}$ around driven GPS trajectories are treated as positives $P$,
while all remaining pixels are unlabeled $U$. The formulation also suits the data as the traces come from unknown vehicles, so reward-recovery methods that model a specific agent cannot be applied, whereas a positive-unlabelled treatment requires only that traversed terrain was passable.

The nnPU estimator is defined over positive and negative risks, each an expectation of a per-pixel classification loss; we instantiate the loss as binary cross-entropy since the path-logits head predicts a per-pixel probability. The PU formulation prevents unlabeled pixels from being treated as negatives. For each pixel $\mathbf{x}$, the predicted path likelihood is $\hat{p}=\sigma(f(\mathbf{x}))$. The positive and negative BCE terms are
\begin{equation}
\ell^+(\mathbf{x})=-\log\hat{p},\qquad
\ell^-(\mathbf{x})=-\log(1-\hat{p}).
\label{eq:bce_terms}
\end{equation}
The corresponding risks are
\begin{equation}
\begin{aligned}
R^+_{P} &= \frac{1}{|P|}\sum_{i\in P}\ell^+_i, &
R^-_{P} &= \frac{1}{|P|}\sum_{i\in P}\ell^-_i, &
R^-_{U} &= \frac{1}{|U|}\sum_{i\in U}\ell^-_i.
\end{aligned}
\label{eq:pu_risks}
\end{equation}
With $\Delta=R^-_{U}-\pi R^-_{P}$ presents an unbiased estimate of the negative risk with $\pi$ the corridor prior probability estimated from the training set; corrects for the fact that true positives are present among unlabelled pixels.  The nnPU loss~\cite{kiryo2017positive} is
\begin{equation}
\mathcal{L}_{\text{path}} =
\begin{cases}
\pi R^+_{P} - \Delta, & \Delta < 0,\\
\pi R^+_{P} + \Delta, & \text{otherwise}.
\end{cases}
\label{eq:nnpu}
\end{equation}

Two regularizers further constrain the spatial structure of the predicted corridor. A total-variation term encourages neighboring path probabilities to vary smoothly, while a mass penalty discourages overly broad positive predictions:
\begin{equation}
\mathcal{L}_{\mathrm{TV}} = \tfrac{1}{N}\sum_{i,j}\!\left(|\Delta_x \hat{p}_{i,j}| + |\Delta_y \hat{p}_{i,j}|\right),
\quad
\mathcal{L}_{\mathrm{mass}} = \tfrac{1}{N}\sum_i \hat{p}_i.
\label{eq:tv_mass}
\end{equation}
$N = HW$, $H, W$ corresponds to height and width.
\subsubsection{Auxiliary Traversability Supervision}

Path supervision alone does not constrain traversability $\hat{T}$ outside observed trajectories. We therefore use OSM road/trail geometries and aggregated historical trajectories as weak positive signals. These are treated as soft supervisory anchors rather than ground-truth traversability labels, since they do not provide dense or platform-specific measurements.

For mapped roads and trails, we use a target of $1$ to encourage high traversability:
\begin{equation}
\mathcal{L}_{\mathrm{track}} =
\frac{\sum_i m_i^{\mathrm{osm}}(\hat{T}_i-\tau_{\mathrm{trk}})^2}
{\sum_i m^{\mathrm{osm}}_i}.
\label{loss_track}
\end{equation}
For historical trajectories, we use the softer target $\tau_{\mathrm{tj}}=0.75$ to provide positive guidance while allowing greater uncertainty:
\begin{equation}
\mathcal{L}_{\mathrm{traj}} =
\frac{\sum_i m^{\mathrm{traj}}_i(\hat{T}_i-\tau_{\mathrm{tj}})^2}
{\sum_i m^{\mathrm{traj}}_i}
\label{loss_traj}
\end{equation}
Here $m^{\mathrm{osm}}$ and $m^{\mathrm{traj}}$ are binary masks of the rasterised OSM geometries and aggregated trajectories, and $\hat{T}_i\in[0,1]$ is the predicted traversability, while $\tau_{\mathrm{trk}}$ and $\tau_{\mathrm{tj}}$ serve as relative supervisory anchors rather than universally calibrated traversability scores.

\subsubsection{LiDAR-derived priors}
The auxiliary terms provide positive traversability evidence but do not explicitly penalize hazardous terrain. We therefore derive three supervised priors from LiDAR that constrain $\hat{T}$ using complementary geometric and surface cues.

\textbf{Intensity prior.} LiDAR intensity provides a surface cue that distinguishes likely traversable surfaces from vegetation and water. With $I$ normalized to $[0,1]$ and $\tilde{I}=1-I$ used as the network input, we regularize $\hat{T}$ toward $I$:
\begin{equation}
\mathcal{L}_{\text{int}} = \tfrac{1}{N}\sum_{i}(\hat{T}_i - I_i)^{2}
\label{eq:intensity_prior}
\end{equation}

\textbf{Slope prior.} Slope provides a direct geometric constraint on vehicle traversability. We map slope $s$ to a bounded soft target using a Gaussian decay:
\begin{equation}
    \tau_{\mathrm{sl}} = \exp\!\Bigl(-\tfrac{s^{2}}{2\sigma_{s}^{2}}\Bigr), \qquad
    \mathcal{L}_{\text{slope}} = \tfrac{1}{N}\sum_{i}(\hat{T}_i - \tau_{\mathrm{sl,i}})^{2}
    \label{eq:slope_prior}
\end{equation}
where $\sigma_s=0.445$, corresponding to a traversability crossover near $30^\circ$.

\textbf{Height-gradient prior.} Height gradients capture abrupt elevation changes such as ditches and drop-offs that slope alone may not characterize. The gradient magnitude $g$ is normalized by its 95th percentile $g_{95}$ and mapped to a soft traversability target:
\begin{gather}
    \tilde{g} = \mathrm{clamp}\!\Bigl(\tfrac{g}{g_{95}},0,1\Bigr), \qquad
    \tau_{\mathrm{ht}} = \exp\!\Bigl(-\tfrac{\tilde{g}^{2}}{2\sigma_{h}^{2}}\Bigr) \notag \\
    \mathcal{L}_{\text{height}} = \tfrac{1}{N}\sum_{i}(\hat{T}_i - \tau_{\mathrm{ht,i}})^{2}
\label{eq:height_prior}
\end{gather}
Together, these priors provide complementary surface, slope, and elevation-discontinuity cues.

\subsubsection{Total objective}
All auxiliary terms are gated by $\alpha_{\text{aux}}(t) \in [0,1]$, linearly 
ramped over $E_{\text{warm}}{=}3$ epochs so that the path-corridor signal 
dominates early optimisation before weaker supervision is introduced, 
substantially improving training stability:
\begin{equation}
\mathcal{L} = \sum_{k \in \mathcal{P}} \lambda_k \mathcal{L}_k
  \;+\; \alpha_{\text{aux}}(t) \sum_{k \in \mathcal{A}} \lambda_k \mathcal{L}_k
\label{eq:total_loss}
\end{equation}
where $\mathcal{P}=\{\text{path},\text{mass},\text{tv}\}$ are the path-supervision
terms and $\mathcal{A}=\{\text{trk},\text{tj},\text{int},\text{sl},\text{ht}\}$ the
anchor terms, gated by $\alpha_{\text{aux}}(t)$. The path corridor weight $\lambda_{\mathrm{p}} = 1.0$ serves as the primary
supervision signal, with auxiliary and prior terms weighted an order of
magnitude smaller ($\lambda \leq 0.04$). One weight set is used for every scene and experiment. The anchors are weighted weakly because they need not agree with one another. Treating them as labels would drive $\hat{T}$ toward their mean where they conflict, whereas at $\lambda \leq 0.04$ they bias feature selection and leave the corridor loss to resolve those cells.

% \subsection{Training details}
\section{Evaluation and Ablation Study}
The 299 scenes are partitioned at the scene level where 20 are  held out for testing, the remaining 279 split into training and validation with $0\%$ footprint overlap between splits. We trained on an RTX 4090 with 24\,GB vRAM using 7533 training and 1716 validation trajectories; the model converged to a total loss of $0.065$ on the training set and $0.076$ on validation.

\subsection{Benchmark}
\label{sec:benchmark}
Traversability depends on robot embodiment, terrain, and operator preference,
none of which admits a canonical label. We therefore evaluate a costmap by how
closely the paths it induces reproduce driven reference trajectories. This
measures agreement with demonstrated practice rather than optimality, since
driven routes reflect habit and familiarity as well as terrain.

\paragraph{Baselines.}
We compare against a continuous-valued height map (elevation normalised per
scene), Trailblazer~\cite{viswanath2025trailblazerlearningoffroadcostmaps},
OVerSeec~\cite{rana2026overseec}\footnote{Queried with the prompt: \emph{``I
prefer roads and established trails, with grasslands and open areas as the next
preferred terrain. Areas including quarries, water bodies, and dense vegetation
are lethal.''}}, and a hand-designed
geometric costmap. The last combines slope~\cite{Horn1981HillShading}, terrain
ruggedness~\cite{riley1999terrain}, step height, an intensity-derived
vegetation cue computed directly from the raw aerial LiDAR returns, together
with the same OSM semantic priors. The features are blended by a weighted
sum whose weights are tuned \emph{per site} by exhaustive sweep.

\paragraph{Protocol and metrics.}
For each costmap we generate $K{=}4$ distinct candidate \emph{paths} with Yen's
algorithm~\cite{yen1971finding} using A$^\ast$ as the underlying search, and
retain the one with the smallest discrete Fr\'{e}chet
distance~\cite{eiter1994computing}, $d_F$, to the driven path. Yen's algorithm
supplies the diversity a single deterministic A$^\ast$ query cannot, and
selecting the closest candidate is deliberate. Several routes are typically
viable off-road, so we ask whether the driven route is among those the map ranks
highly rather than penalising a map for proposing an equally viable alternative.
The same $K$ and selection rule apply to every method. Let $C$ denote the
costmap (higher $=$ costlier) and $P, D$ the planner and driven paths as pixel
sequences. We report three axes, path shape, cost alignment, and length.
\begin{equation}
    d_F(P,D), \qquad
    r_c = \frac{\bar{c}_P}{\bar{c}_D}, \qquad
    r_d = \frac{L_P}{L_D},
    \label{eq:metrics}
\end{equation}
\begin{table}[t]
    \centering
    \small
    \setlength{\tabcolsep}{5pt}
    \begin{tabular}{lccc}
        \toprule
        \textbf{Costmap}
            & $d_F$ (m) $\downarrow$
            & $r_c$ ($\!\to\!1$)
            & $r_d$ ($\!\to\!1$) \\
        \midrule
        Height map
            & $108.4 \pm 18.2$ & $0.85 \pm 0.11$ & $0.91 \pm 0.06$ \\
        Trailblazer~\cite{viswanath2025trailblazerlearningoffroadcostmaps}
            & $91.6 \pm 22.7$  & $0.71 \pm 0.09$ & $0.81 \pm 0.08$ \\
        OVerSeeC~\cite{rana2026overseec}
            & $77.5 \pm 20.1$  & $0.23 \pm 0.14$ & $0.92 \pm 0.06$ \\
        Geometric$^\dagger$~\cite{Horn1981HillShading,riley1999terrain}
            & $68.4 \pm 17.8$ & $0.78 \pm 0.09$ & $0.86 \pm 0.09$  \\
        \textbf{Ours}
            & $\mathbf{61.3 \pm 14.3}$ & $0.93 \pm 0.05$ & $0.89 \pm 0.05$ \\
        \bottomrule
    \end{tabular}
    \caption{Costmap benchmark on human trajectories (mean $\pm$ std,
    $n{=}356$); best $d_F$ in bold. $^\dagger$Weights tuned \emph{per site};
    all other methods are without per-site adjustments.} 

    \label{tab:benchmark}

\end{table}
where $\bar{c}_X=|X|^{-1}\!\sum_{(i,j)\in X}\!C(i,j)$ and
$L_X=\sum_k \lVert x_{k+1}-x_k\rVert$, with perfect alignment at $d_F{=}0$,

As $r_c$ and $r_d$ are evaluated within each costmap,
they are diagnostics rather than cross-method scores. A spatially uniform map
would attain $r_c{=}1$ by construction while being uninformative. Only $d_F$,
measured in metres against human driving and independent of the map being
scored, ranks methods on its own. Both ratios are computed on the
$d_F$-selected candidate.

\paragraph{Results.}
Table~\ref{tab:benchmark} reports means and standard deviations over $n{=}356$
trajectories (average length $412$\,m) from 20 held-out test scenes, including
TartanDrive 2.0~\cite{sivaprakasam2024tartandrive}. TartanDrive 2.0 trajectories are cropped to goal-directed segments before evaluation.

Ours attains the lowest $d_F$ at $61.3$\,m, $10\%$ below the strongest baseline
($68.4$\,m, tuned per site) and wins on $62.5\%$ of trajectories individually
(paired Wilcoxon, $p{<}10^{-3}$). The geometric baseline's tuned weights diverge
across sites, so no single hand-tuned
parameters transfers. Ours is also closest to cost parity ($r_c{=}0.93$):
the demonstrated corridor is near-optimal under the learned cost rather than
merely reachable. The two counter-cases show why the ratios cannot rank methods
alone as OVerSeeC's favourable $r_d{=}0.92$ accompanies $r_c{=}0.23$, a shortcut
through terrain the human avoided, while the height map attains competitive
ratios with the worst $d_F$. All $r_d$ fall below unity ($0.81$--$0.92$) as
A$^\ast$ takes near-geodesic routes whereas drivers satisfice; ours ($0.89$) is
mid-range.
% Together, these results show that our costmap captures human traversability preference more accurately than the compared baselines, both geometrically and in induced planner cost.

\subsection{Ablation Study}
To analyse the contribution of each loss group, we train five variants of our
model:
\begin{itemize}
    \item \textbf{Variant A:} Path corridor only ($\mathcal{L}_{\text{path}} + \mathcal{L}_{\text{mass}} + \mathcal{L}_{\text{TV}}$)
    \item \textbf{Variant B:} A + auxiliary terms ($\mathcal{L}_{\text{track}} + \mathcal{L}_{\text{traj}}$)
    \item \textbf{Variant C:} A + LiDAR priors ($\mathcal{L}_{\text{int}} + \mathcal{L}_{\text{slope}} + \mathcal{L}_{\text{height}}$)
    \item \textbf{Variant D:} Full loss (all 8 terms)
    \item \textbf{Variant E:} Single head. Corridor terms applied to $\hat{T}$
    directly at $\lambda_{\mathrm{p}}{=}1.0$, anchors and priors as in D, no
    goal conditioning.
\end{itemize}
All variants share identical training data and hyperparameters. We evaluate on
the 73 held-out trajectories that admit two routes between the same endpoints,
one a \emph{dense corridor} with multiple overlapping traces and one a
\emph{sparse alternative} with few. Where many traces overlap, their density
provides weak evidence of preference beyond mere traversability, so dense
corridors are a stronger reference than any single trace. Both routes are
geometrically accessible, and a map that discriminates terrain should separate
them. A larger gap $\Delta$ therefore indicates that the map distinguishes
terrain vehicles actually use from terrain that is merely passable.
\begin{table}[h]
    \centering
    \small
    \begin{tabular}{lccccc}
        \toprule
        & A & B & C & D & E \\
        \midrule
        Dense corridor     & 0.561 & 0.716 & 0.905 & 0.980 & 0.926 \\
        Sparse alternative & 0.542 & 0.681 & 0.781 & 0.780 & 0.832 \\
        \midrule
        $\Delta$           & 0.019 & 0.035 & 0.124 & \textbf{0.200} & 0.094 \\
        \bottomrule
    \end{tabular}
    \caption{Mean traversability along \emph{dense corridors} vs.\ \emph{sparse
    alternatives} per ablation variant, on 73 paired held-out trajectories.}
    \label{tab:ablation}

\end{table}
\begin{figure}
    \centering
    \includegraphics[width=\linewidth]{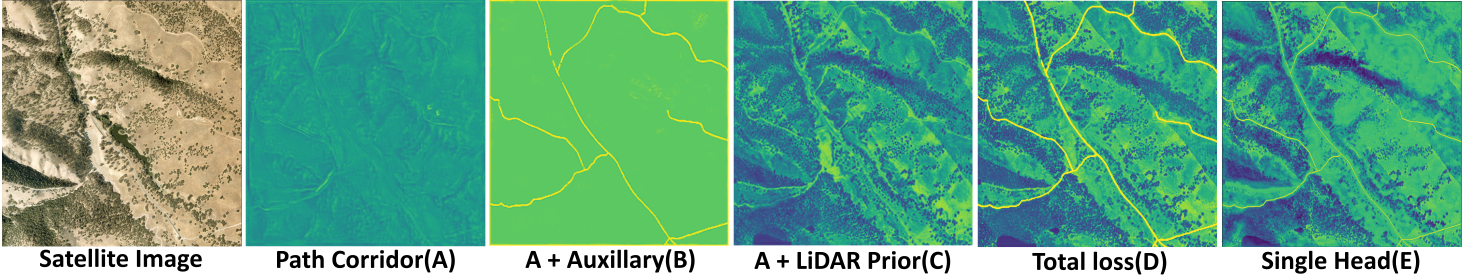}
    \caption{Ablation traversability maps on a held-out desert scene. Satellite image,
then Variants A to E of Table~II. Path supervision alone (A) leaves
traversability near-uniform outside the corridor. LiDAR priors (C) and the full
loss (D) recover terrain structure. The single-head variant (E) expresses the corridors weakly and flattens the
separation between them and the surrounding terrain.}
    \label{fig:ablation}

\end{figure}
Variant~A produces near-identical scores for both route types
($\Delta{=}0.019$), confirming that path supervision alone is insufficient for
terrain-aware costmaps. Auxiliary terms (B, $\Delta{=}0.035$) and LiDAR priors
(C, $\Delta{=}0.124$) each improve discrimination, and neither reaches the full
loss ($\Delta{=}0.200$), so the two are complementary rather than redundant. The
manner of the improvement is also informative. From C to D the sparse
alternatives are unchanged ($0.781 \to 0.780$) while the dense corridors rise
($0.905 \to 0.980$), so the full loss widens the gap by recognising driven
terrain rather than by suppressing terrain that is merely accessible.
Variant~E attains the highest sparse-alternative score and a gap below C. With a
single head the corridor supervision and the geometric anchors compete within
one output, so the terrain field is smoothed toward the middle and the corridors
are only weakly expressed. Separating the heads lets the corridor loss act at
full weight without that competition, which is why D resolves the trail network
while retaining terrain structure.
Qualitative comparisons in Figure~\ref{fig:ablation} corroborate these findings.

\section{Experimental Setup}

\begin{figure}[t]
    \centering
    \includegraphics[width=0.55\linewidth, height = 4 cm]{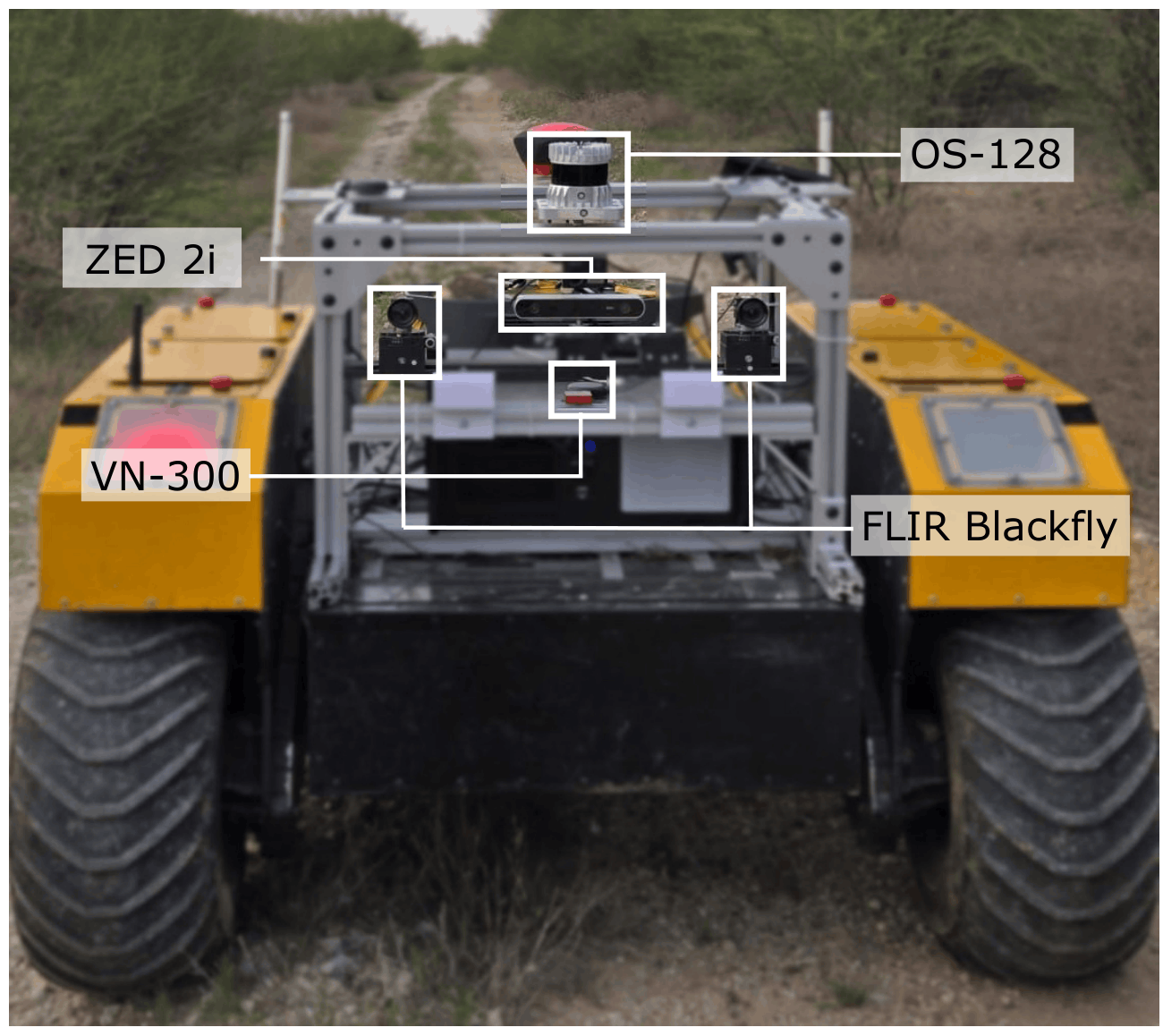}
    \caption{Sensor suite integrated on the Warthog platform}
    \label{fig:sensors}

\end{figure}

Offline benchmarking alone cannot capture the system-level effects 
that dominate real-world off-road deployment. Hardware integration, 
inter-module coordination, and edge-case robustness. We therefore 
evaluate on a physical platform, a Clearpath Warthog 
equipped with the multi-modal sensor suite 
Fig.~\ref{fig:sensors}. Field experiments were conducted at speeds 
up to $2\,\mathrm{m/s}$.

Our algorithm serves as \textbf{Long Range Planner (LRP)}, generating global waypoint sequences from the robot's current pose to a user-specified goal. These waypoints are consumed by ARL's Phoenix autonomy stack, which handles short-range planning and control: Direct LiDAR-Inertial Odometry~\cite{chen2022dlio} for state estimation, and Model Predictive Path Integral (MPPI) control~\cite{williams2017information} as the \textbf{Short Range Planner (SRP)}, operating over local costmaps from TerrainNet~\cite{MengHLLWSLYCDOF23} under the Warthog's kinematic-bicycle constraints. The LRP thus provides the global plan while the SRP handles reactive obstacle avoidance from live onboard perception.

\subsection{Global Planner}

For downstream planning, the traversability map is transformed into a cost representation via an inverse mapping $C = 1 - \hat{T}$, aligning high traversability with low planning cost. Given a goal, the LRP queries the resulting cost map to compute $K$-shortest paths using Yen's algorithm. Search operates on the discrete $x$--$y$ grid, so the LRP output is a geometric reference and dynamic feasibility is delegated to the SRP.

The selected path is transformed from the map frame to UTM for global consistency and then discretized into waypoints. The resulting waypoint sequence is streamed to the SRP for reactive execution.

\section{Experiments and Insights}

Through our experiments, we address the following questions:
\begin{enumerate}
    \item Does the learned costmap sufficiently reflect terrain characteristics derived from satellite, LiDAR, and map-based inputs?
    \item Does the long-range planner yield trajectories that are measurably safer, shorter, or more reliable than those produced by local-planner-only autonomy?
    \item Is the learned costmap expressive enough to yield multiple distinct yet feasible trajectories between the same start–goal pair?
    \item How does the planner handle outdated costmaps when encountering unseen obstacles or terrain changes?
\end{enumerate}
 
\begin{figure}[t]
    \centering
    \includegraphics[width=0.9\linewidth, height = 5.5 cm]{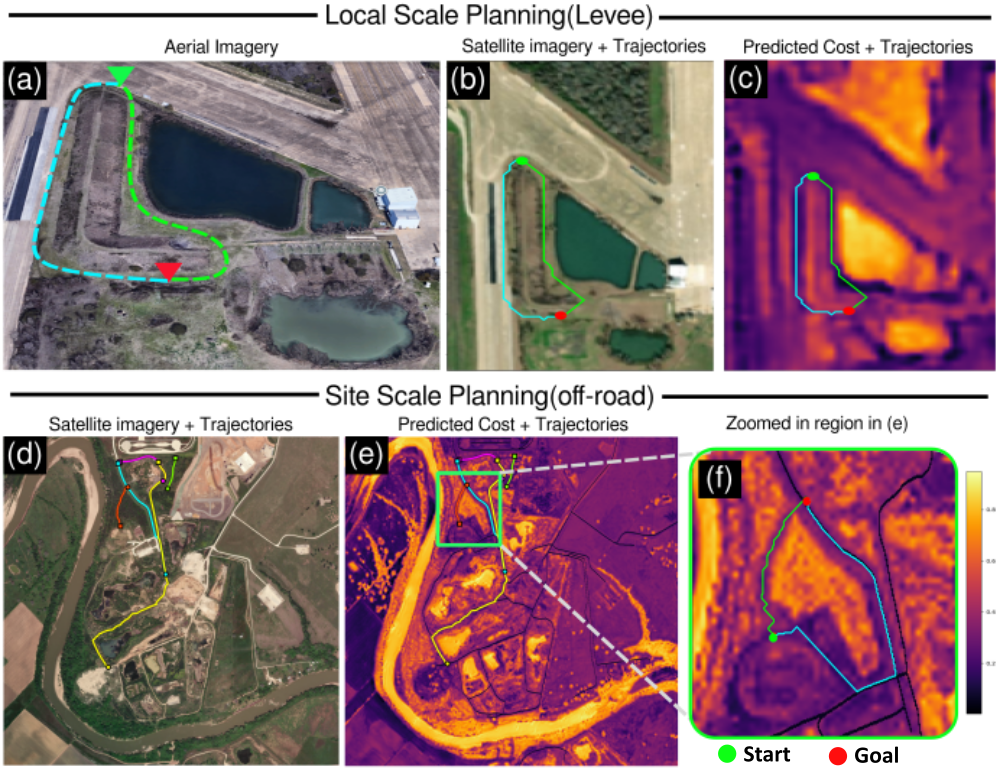}
    \caption{%
        Overview of the Levee test site (a, b), showing the 
        $250\,\mathrm{m}$ levee embankment, adjacent lake, and runway 
        segment, with its learned costmap (c) used to evaluate slope 
        and water-body avoidance. Overview of the off-road site (d) 
        and its costmap (e), with routes used for experiments. (f) Zoomed view 
        demonstrating the model's ability to identify traversable 
        corridors through dense vegetation.%
    }
    \label{fig:levee_test_site}

\end{figure}

Real-world experiments were conducted at two sites: a \textit{Levee
site} and an \textit{off-road site} at Texas A\&M's RELLIS Campus. Neither site appears in the training corpus, and the same model weights are used at both, so both deployments are zero-shot with respect to geography.
The Levee site features a 250\,m levee embankment, an adjacent
lake, and a segment of a runway (Figure~\ref{fig:levee_test_site}a, ~\ref{fig:levee_test_site}b, ~\ref{fig:levee_test_site}c), and was selected to evaluate algorithm performance around slopes and water bodies. The off-road site spans more than $9\,\text{km}^{2}$ and encompasses diverse terrain types like grassland, quarry, trails, lakes, and deciduous forest (Figure~\ref{fig:levee_test_site}d, ~\ref{fig:levee_test_site}e) enabling comprehensive evaluation across varied conditions.

\subsection*{\textbf{Does the costmap sufficiently capture terrain features?}}

We evaluated the learned costmap across diverse terrain, uphill slopes 
and water bodies at the Levee site, and dense vegetation and deep 
quarries at the off-road site. At the Levee site 
(Fig.~\ref{fig:levee_test_site}c), the costmap correctly assigned high cost to the embankment slopes and adjacent water despite their grass-like appearance in satellite imagery, guiding the robot through the levelled corridor between them. On Route~1 (Fig.~\ref{fig:lrp_srp_comparison}), the flat region 
beyond the levee was initially occluded from the SRP's local perception, 
causing it to climb the embankment and approach the water dangerously 
close, a failure mode the LRP avoided by routing around the levee with 
safe water-body clearance.

At the off-road site, the LRP followed established trails as low-cost 
corridors while excluding quarries, negative obstacles that are 
difficult to detect from onboard sensors but clearly identifiable from 
overhead LiDAR geometry. The costmap also resolved traversable 
corridors through wooded areas (Fig.~\ref{fig:levee_test_site}f).

We note, however, that the costmap tends to be over-conservative in 
transitional zones such as grass-to-brush boundaries and drainage 
edges, contributing to the modest path-length overhead discussed next. 
This is consistent with our PU-supervised loss, which lacks explicit 
negative examples of ``traversable but unusual'' terrain.

\subsection*{\textbf{Does the global planner outperform local-planner-only autonomy?}}

\begin{figure}[t]
    \centering
    \includegraphics[width=0.9\linewidth, height = 6.5 cm]%
        {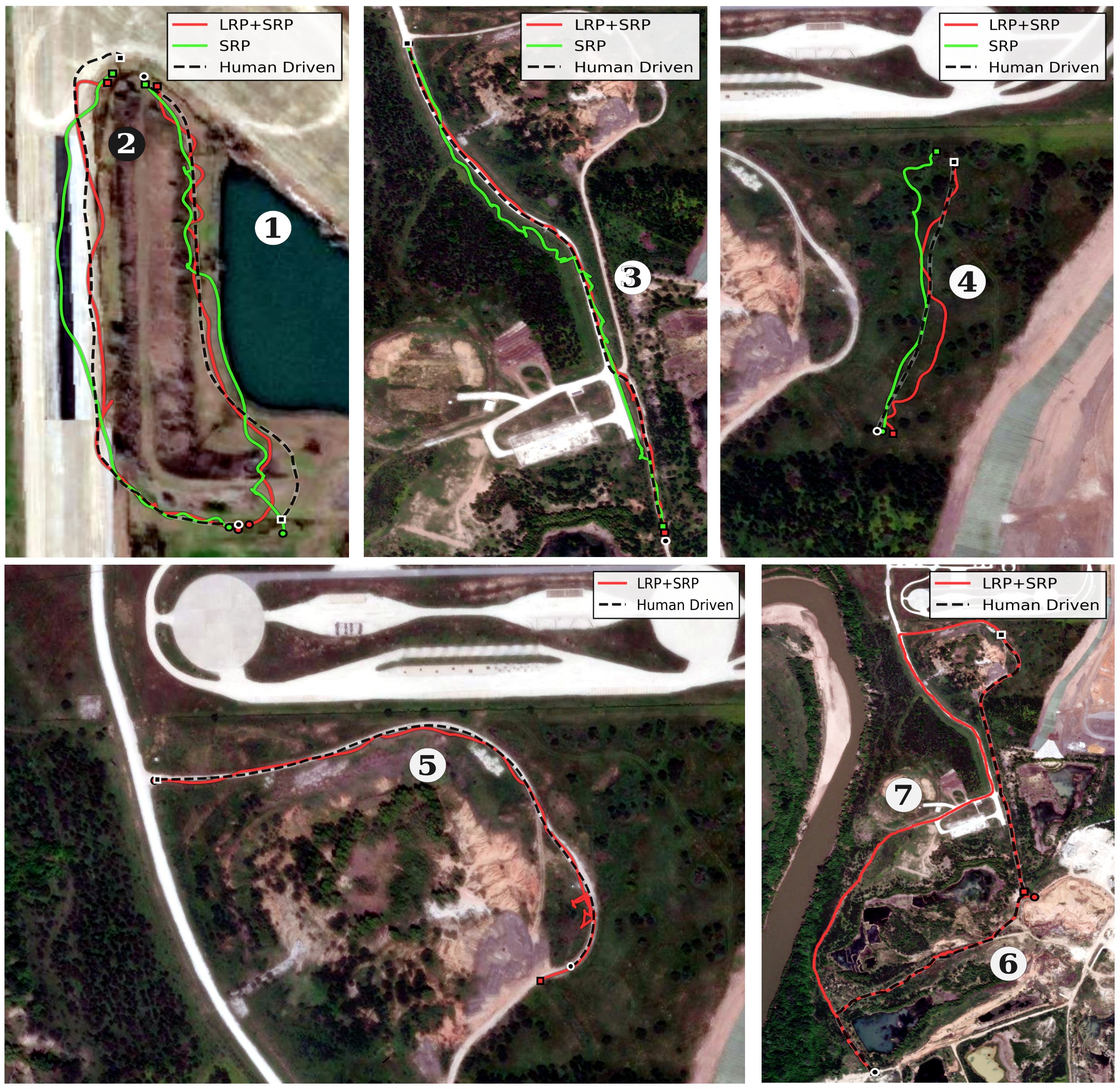}
    \caption{%
        Trajectory comparison across seven routes, illustrating the improved path quality and reduced deviation of \textcolor{red}{LRP}  relative to \textcolor{green}{SRP}-only autonomy. Route 1 and 2 are at the Levee site and the rest at the off-road test site.
    }
    \label{fig:lrp_srp_comparison}

\end{figure}

We use human teleoperation tracks as the baseline, in which an operator manually navigates the robot along a preferred route while avoiding obstacles. We then compare full-stack autonomy (LRP) against SRP-only navigation across identical routes. SRP-only runs were conducted on a limited subset of routes (Table~\ref{tab:quantitative_metrics}), as SRP relies on a pure-pursuit global guidance mechanism that poses unacceptable safety risks near hazards such as quarry edges or long horizon missions. Quantitative results are presented in Table~\ref{tab:quantitative_metrics}, and qualitative trajectory comparisons are shown in
Figure~\ref{fig:lrp_srp_comparison}.

\begin{table}[h]
    \centering
    \fontsize{7pt}{9pt}\selectfont
    \setlength{\tabcolsep}{3pt}
    \begin{tabular}{|c|c|c|c|c|c|}
        \hline
        \makecell{Route} &
        \makecell{Distance\\Human (m)} &
        \makecell{Distance\\LRP (m)} &
        \makecell{Distance\\SRP (m)} &
        \makecell{Intervention\\LRP} &
        \makecell{Intervention\\SRP}  \\
        \hline
        1 & 320  & 335  & 360 & 0 & 3 \\
        \hline
        2 & 379  & 386  & 392 & 1 & 2 \\
        \hline
        3 & 765  & 810  & 901 & 1 & 8 \\
        \hline
        4 & 195  & 220  & 236 & 0 & 1 \\
        \hline
        5 & 393  & 425  & --- & 2 & ---  \\
        \hline
        6 & 1680 & 1714 & --- & 3 & --- \\
        \hline
        7 & 1680 & 1956 & --- & 1 & --- \\
        \hline
    \end{tabular}
    \caption{%
        Quantitative comparison of human teleoperation, LRP autonomy, and SRP-only autonomy across seven routes. Dashes (---) indicate routes where SRP-only runs were not performed due to safety constraints.
    }
    \label{tab:quantitative_metrics}

\end{table}

On the four routes where both were run, LRP achieved trajectories closest to the
human baseline, with an average distance overhead of \textbf{5.5}\% compared to \textbf{13.9}\% for SRP-only. Operator interventions triggered by obstacle encounters or large deviations were \textbf{85}\% lower for LRP on these routes (2 vs.\ 14), averaging one intervention per \textbf{876}\,m against one per \textbf{135}\,m for SRP. Across all seven routes, including the three long-horizon runs SRP could not attempt, LRP overhead averaged $7.4\%$.

\emph{Where did the LRP fall short of the human?} 
\begin{itemize}
   \item \textbf{Tracking-induced deviation from the planned path.} 
    The LRP-planned routes remain geometrically close to the human trajectories (Fig.~\ref{fig:lrp_srp_comparison}), however during downstream execution, small lateral deviations accumulate due to MPPI tracking error, the Warthog's kinematic constraints, and reactive local re-planning around short-range hazards.
    
    \item \textbf{Model conservatism in ambiguous vegetation.} 
    A smaller part of the gap comes from the learned costmap itself, as noted in Q1. In ambiguous vegetation, most clearly on Route~4, LRP occasionally routes around lightly-wooded corridors. A human operator would cut through them directly.
\end{itemize}
 
\subsection*{\textbf{Can the learned costmap produce multiple feasible paths?}}

Querying the planner with identical start--goal pairs yields distinct
alternatives at both sites: paths of comparable length circumnavigating the
Levee embankment via separate corridors (Routes 1 \& 2), and substantially
different long-horizon routes to the same goal at the off-road site
(Routes 6 \& 7). The secondary off-road route was longer in absolute distance
but only $5.9\%$ higher in cost per distance, correctly reflecting its traversal
of marginally less-preferred terrain (a mixed-vegetation corridor rather than an
established trail). Route diversity of this kind depends on the costmap as well
as the planner: cost values must be distributed finely enough for multiple
trajectories to remain competitive.

\subsection*{\textbf{How does the planner handle outdated costmaps with unseen obstacles?}}

When the platform encounters an obstacle absent from the global costmap during execution, the SRP detects the blockage and halts the robot. This event triggers a replanning request to the LRP: a high-cost region is injected at the obstacle's estimated location, and an alternative route is computed from the current position to the goal. During testing on Route~6, the platform encountered an unmarked sand heap (Figure~\ref{fig:lrp_srp_comparison}); the LRP successfully replanned around it and reached the goal without further intervention.

\section{Conclusion and Future Work}

In this work, we presented a dual-stream multi-modal framework that learns continuous traversability maps from satellite imagery, aerial LiDAR, and OpenStreetMap priors, supervised directly by sparse human trajectories and geometric priors. We demonstrate that the learned costmaps enable globally-aware, terrain-sensitive navigation that measurably outperforms local-planner-only autonomy in both path efficiency and operator intervention rate. However, the LRP--SRP interface remains sensitive to lookahead distance and handoff timing, and static costmaps cannot account for transient obstacles or seasonal terrain change. Future work includes online costmap refinement fusing the global prior with onboard perception, and tighter LRP--SRP co-design which are prerequisites for reliable high-speed off-road autonomy.

\section{Acknowledgments}
We thank Felix Sanchez, Phil Osteen, and Eli Lancaster for their help during field testing, as well as the members of the Unmanned Systems Lab for their valuable discussions and feedback. We gratefully acknowledge the support of the Texas A\&M RELLIS Campus administration for providing access to the test sites.  
\bibliographystyle{IEEEtran}
\bibliography{IEEEabrv,Reference}

@article{https://doi.org/10.1002/rob.20168,
author = {Howard, Andrew and Turmon, Michael and Matthies, Larry and Tang, Benyang and Angelova, Anelia and Mjolsness, Eric},
title = {Towards learned traversability for robot navigation: From underfoot to the far field},
journal = {Journal of Field Robotics},
volume = {23},
number = {11-12},
pages = {1005-1017},
doi = {https://doi.org/10.1002/rob.20168},
url = {https://onlinelibrary.wiley.com/doi/abs/10.1002/rob.20168},
eprint = {https://onlinelibrary.wiley.com/doi/pdf/10.1002/rob.20168},
year = {2006}
}

@article{yen1971finding,
  title={Finding the k shortest loopless paths in a network},
  author={Yen, Jin Y},
  journal={Management Science},
  volume={17},
  number={11},
  pages={712--716},
  year={1971},
  publisher={INFORMS}
}

@article{Hart1968,
  doi = {10.1109/tssc.1968.300136},
  url = {https://doi.org/10.1109/tssc.1968.300136},
  year = {1968},
  publisher = {Institute of Electrical and Electronics Engineers ({IEEE})},
  volume = {4},
  number = {2},
  pages = {100--107},
  author = {Peter Hart and Nils Nilsson and Bertram Raphael},
  title = {A Formal Basis for the Heuristic Determination of Minimum Cost Paths},
  journal = {{IEEE} Transactions on Systems Science and Cybernetics}
}

@inproceedings{MengHLLWSLYCDOF23,
  author = {Meng, Xiangyun and Hatch, Nathan and Lambert, Alexander and Li, Anqi and Wagener, Nolan and Schmittle, Matthew and Lee, Joonho and Yuan, Wentao and Chen, Zoey Qiuyu and Deng, Samuel and Okopal, Greg and Fox, Dieter and Boots, Byron and Shaban, Amirreza},
  booktitle = {Robotics: Science and Systems},
  ee = {https://doi.org/10.15607/RSS.2023.XIX.103},
  isbn = {978-0-9923747-9-2},
  title = {TerrainNet: Visual Modeling of Complex Terrain for High-speed, Off-road Navigation.},
  url = {http://dblp.uni-trier.de/db/conf/rss/rss2023.html#MengHLLWSLYCDOF23},
  year = 2023
}

@inproceedings{williams2017information,
  title={Information theoretic MPC for model-based reinforcement learning},
  author={Williams, Grady and Wagener, Nolan and Goldfain, Brian and Drews, Paul and Rehg, James M and Boots, Byron and Theodorou, Evangelos A},
  booktitle={2017 IEEE International Conference on Robotics and Automation (ICRA)},
  pages={1714--1721},
  year={2017},
  organization={IEEE},
  doi={10.1109/ICRA.2017.7989202},
  url={https://ieee.org}
}

@inproceedings{10.5555/3504035.3504518,
author = {Perez, Ethan and Strub, Florian and de Vries, Harm and Dumoulin, Vincent and Courville, Aaron},
title = {FiLM: visual reasoning with a general conditioning layer},
year = {2018},
isbn = {978-1-57735-800-8},
publisher = {AAAI Press},
booktitle = {Proceedings of the Thirty-Second AAAI Conference on Artificial Intelligence},
articleno = {483},
numpages = {10},
location = {New Orleans, Louisiana, USA},
series = {AAAI'18/IAAI'18/EAAI'18}
}

@article{simeoni2025dinov3,
  title={DINOv3},
  author={Sim{\'eoni}, Oriane and Vo, Huy V. and Seitzer, Maximilian and Baldassarre, Federico and Oquab, Maxime and Jose, Cijo and Khalidov, Vasil and Szafraniec, Marc and Yi, Seung Eun and Ramamonjisoa, Micha{\"e}l and Massa, Francisco and Haziza, Daniel and Wehrstedt, Luca and Wang, Jianyuan and Darcet, Timoth{\'e}e and Moutakanni, Th{\"e}o and Sentana, Leonel and Roberts, Claire and Vedaldi, Andrea and Tolan, Jamie and Brandt, John and Couprie, Camille and Mairal, Julien and J{\'e}gou, Herv{\'e} and Labatut, Patrick and Bojanowski, Piotr},
  journal={CoRR},
  volume={abs/2508.10104},
  year={2025},
  month={August},
  url={https://doi.org/10.48550/arXiv.2508.10104}
}

@misc{usgs_lidar_2024,
  author = {{U.S. Geological Survey}},
  title = {3D Elevation Program ({3DEP}) - {Lidar} {Point} {Cloud}},
  year = {2024},
  url = {https://apps.nationalmap.gov/lidar-explorer/},
  note = {Accessed: 2026},
  organization = {U.S. Geological Survey}
}

@misc{OpenStreetMap,
  author = {{OpenStreetMap contributors}},
  title = {{Planet dump retrieved from https://planet.osm.org }},
  year = {2025},
  note = {Accessed: [2026]}
}

@misc{esri_world_imagery,
  author = {{Esri, DigitalGlobe, GeoEye, Earthstar Geographics, CNES/Airbus DS, USDA, USGS, AeroGRID, IGN, and the GIS User Community}},
  title = {World Imagery [Basemap]},
  year = {2026},
  url = {https://www.arcgis.com/home/item.html?id=10df2279f9684e4a9f6a7f08febac2a9},
  note = {Accessed: 2026}
}

@misc{sivaprakasam2024tartandrive,
      title={TartanDrive 2.0: More Modalities and Better Infrastructure to Further Self-Supervised Learning Research in Off-Road Driving Tasks},
      author={Matthew Sivaprakasam and Parv Maheshwari and Mateo Guaman Castro and Samuel Triest and Micah Nye and Steve Willits and Andrew Saba and Wenshan Wang and Sebastian Scherer},
      year={2024},
      eprint={2402.01913},
      archivePrefix={arXiv},
      primaryClass={cs.RO}}

@INPROCEEDINGS{9561251,
  author={Jiang, Peng and Osteen, Philip and Wigness, Maggie and Saripalli, Srikanth},
  booktitle={2021 IEEE International Conference on Robotics and Automation (ICRA)}, 
  title={RELLIS-3D Dataset: Data, Benchmarks and Analysis}, 
  year={2021},
  volume={},
  number={},
  pages={1110-1116},
  doi={10.1109/ICRA48506.2021.9561251}}

@inproceedings{goose-dataset,
    author = {Peter Mortimer and Raphael Hagmanns and Miguel Granero
              and Thorsten Luettel and Janko Petereit and Hans-Joachim Wuensche},
    title = {The GOOSE Dataset for Perception in Unstructured Environments},
    url={https://arxiv.org/abs/2310.16788},
    booktitle={Proceedings of IEEE International Conference on Robotics and Automation (ICRA)},
    year = {2024}
}

@misc{jiang2025gogreatoutdoorsmultimodal,
      title={GO: The Great Outdoors Multimodal Dataset}, 
      author={Peng Jiang and Kasi Viswanath and Akhil Nagariya and George Chustz and Maggie Wigness and Philip Osteen and Timothy Overbye and Christian Ellis and Long Quang and Srikanth Saripalli},
      year={2025},
      eprint={2501.19274},
      archivePrefix={arXiv},
      primaryClass={cs.RO},
      url={https://arxiv.org/abs/2501.19274}, 
}

@inproceedings{RUGD2019IROS,
  author = {Wigness, Maggie and Eum, Sungmin and Rogers, John G and Han, David and Kwon, Heesung},
  title = {A RUGD Dataset for Autonomous Navigation and Visual Perception in Unstructured Outdoor Environments},
  booktitle = {International Conference on Intelligent Robots and Systems (IROS)},
  year = {2019}
}

@INPROCEEDINGS{9812238,  author={Weerakoon, Kasun and Sathyamoorthy, Adarsh Jagan and Patel, Utsav and Manocha, Dinesh},  booktitle={2022 International Conference on Robotics and Automation (ICRA)},   title={TERP: Reliable Planning in Uneven Outdoor Environments using Deep Reinforcement Learning},  volume={},  number={},  pages={9447-9453},  doi={10.1109/ICRA46639.2022.9812238}, year={2022}}

@misc{rana2026overseec,
      title={OVerSeeC: Open-Vocabulary Costmap Generation from Satellite Images and Natural Language},
      author={Rwik Rana and Jesse Quattrociocchi and Dongmyeong Lee and Christian Ellis and Amanda Adkins and Adam Uccello and Garrett Warnell and Joydeep Biswas},
      year={2026},
      eprint={2602.18606},
      archivePrefix={arXiv},
      primaryClass={cs.RO},
      url={https://arxiv.org/abs/2602.18606},
}

@misc{hdif2023,
    title={How Does It Feel? Self-Supervised Costmap Learning for Off-Road Vehicle Traversability},
    author={Guaman Castro, Mateo and Triest, Samuel and
    Wang, Wenshan and Gregory, Jason M. and Sanchez, Felix and Rogers III, John G. and Scherer, Sebastian},
    year={2023},
    Booktitle={2023 International Conference on Robotics and Automation (ICRA)},
    organization={IEEE}
}

@misc{viswanath2025trailblazerlearningoffroadcostmaps,
      title={Trailblazer: Learning offroad costmaps for long range planning}, 
      author={Kasi Viswanath and Felix Sanchez and Timothy Overbye and Jason M. Gregory and Srikanth Saripalli},
      year={2025},
      eprint={2505.09739},
      archivePrefix={arXiv},
      primaryClass={cs.RO},
      url={https://arxiv.org/abs/2505.09739}, 
}

@inproceedings{silver2008high,
  title={High Performance Outdoor Navigation from Overhead Data using Imitation Learning},
  author={Silver, David and Bagnell, J. Andrew and Stentz, Anthony},
  booktitle={Robotics: Science and Systems IV},
  year={2008},
  editor={Brock, Oliver and Trinkle, Jeff and Ramos, Fabio},
  publisher={MIT Press},
  url={https://www.roboticsproceedings.org/rss04/p34.pdf}
}

@article{wulfmeier2017large,
  title={Large-scale cost function learning for path planning using deep inverse reinforcement learning},
  author={Wulfmeier, Markus and Rao, Dushyant and Wang, Dominic Zeng and Ondruska, Peter and Posner, Ingmar},
  journal={The International Journal of Robotics Research},
  volume={36},
  number={10},
  pages={1073--1087},
  year={2017},
  publisher={SAGE Publications Sage UK: London, England}
}

@InProceedings{pmlr-v139-yonetani21a,
  title =      {Path Planning using Neural A* Search},
  author    = {Ryo Yonetani and
               Tatsunori Taniai and
               Mohammadamin Barekatain and
               Mai Nishimura and
               Asako Kanezaki},
  booktitle =      {Proceedings of the 38th International Conference on Machine Learning},
  pages =      {12029--12039},
  year =      {2021},
  editor =      {Meila, Marina and Zhang, Tong},
  volume =      {139},
  series =      {Proceedings of Machine Learning Research},
  month =      {18--24 Jul},
  publisher =    {PMLR},
  url =      {http://proceedings.mlr.press/v139/yonetani21a.html},
}

@article{chen2022dlio,
  title={Direct LiDAR-Inertial Odometry: Lightweight LIO with Continuous-Time Motion Correction},
  author={Chen, Kenny and Nemiroff, Ryan and Lopez, Brett T},
  journal={2023 IEEE International Conference on Robotics and Automation (ICRA)},
  year={2023},
  pages={3983-3989},
  doi={10.1109/ICRA48891.2023.10160508}
}

@article{manduchi2005obstacle,
  title={Obstacle detection and terrain classification for autonomous off-road navigation},
  author={Manduchi, Roberto and Castano, Andres and Talukder, Ashit and Matthies, Larry},
  journal={Autonomous Robots},
  volume={18},
  number={1},
  pages={81--102},
  year={2005},
  publisher={Springer}
}

@article{paden2016survey,
  title={A survey of motion planning and control techniques for self-driving urban vehicles},
  author={Paden, Brian and C{\'a}p, Michal and Yong, Sze Zheng and Yershov, Dmitry and Frazzoli, Emilio},
  journal={IEEE Transactions on Intelligent Vehicles},
  volume={1},
  number={1},
  pages={33--55},
  year={2016},
  publisher={IEEE},
  doi={10.1109/TIV.2016.2578706},
  url={https://arxiv.org/abs/1604.07446}
}

@inproceedings{kiryo2017positive,
  title     = {Positive-Unlabeled Learning with Non-Negative Risk Estimator},
  author    = {Kiryo, Ryuichi and Niu, Gang and du Plessis, Marius and Sugiyama, Masashi},
  booktitle = {Advances in Neural Information Processing Systems (NeurIPS)},
  year      = {2017},
}

@inproceedings{eiter1994computing,
  title={Computing discrete Fréchet distance},
  author={Eiter, Thomas and Mannila, Heikki},
  booktitle={Proceedings of the Second International Conference on Computer Vision, Pattern Recognition and Image Processing},
  pages={313--316},
  year={1994},
  organization={IEEE}
}

@inproceedings{Montemerlo2006WinningTD,
  title={Winning the DARPA Grand Challenge with an AI Robot},
  author={Michael Montemerlo and Sebastian Thrun and Hendrik Dahlkamp and David Stavens and Sven Strohband},
  booktitle={AAAI},
  year={2006}
}

@article{riley1999terrain,
  title={A terrain ruggedness index that quantifies topographic heterogeneity},
  author={Riley, Shawn J and DeGloria, Stephen D and Elliot, Robert},
  journal={Intermountain Journal of Sciences},
  volume={5},
  number={1-4},
  pages={23--27},
  year={1999}
}

@article{Horn1981HillShading,
  author = {Horn, Berthold K. P.},
  title = {Hill shading and the reflectance map},
  journal = {Proceedings of the IEEE},
  year = {1981},
  volume = {69},
  number = {1},
  pages = {14--47},
  doi = {10.1109/PROC.1981.11918},
  publisher = {IEEE}
}

@inproceedings{schmittle2025lrn,
  author    = {Schmittle, Matt and Baijal, Rohan and Hatch, Nathan and Scalise, Rosario and Guaman Castro, Mateo and Talia, Sidharth and Khetarpal, Khimya and Boots, Byron and Srinivasa, Siddhartha},
  title     = {Long Range Navigator ({LRN}): Extending Robot Planning Horizons Beyond Metric Maps},
  booktitle = {Proceedings of The 9th Conference on Robot Learning},
  series    = {Proceedings of Machine Learning Research},
  volume    = {305},
  pages     = {3185--3199},
  year      = {2025},
  publisher = {PMLR}
}

@inproceedings{damm2023terrain,
  author       = {Damm, Eric R. and Gregory, Jason M. and Lancaster, Eli S. and Sanchez, Felix A. and Sahu, Daniel M. and Howard, Thomas M.},
  title        = {Terrain-Aware Kinodynamic Planning with Efficiently Adaptive State Lattices for Mobile Robot Navigation in Off-Road Environments},
  booktitle    = {2023 IEEE/RSJ International Conference on Intelligent Robots and Systems (IROS)},
  pages        = {9918--9925},
  year         = {2023},
  organization = {IEEE}
}

@inproceedings{zhu2020offroad,
  author       = {Zhu, Zeyu and Li, Nan and Sun, Ruoyu and Xu, Donghao and Zhao, Huijing},
  title        = {Off-Road Autonomous Vehicles Traversability Analysis and Trajectory Planning Based on Deep Inverse Reinforcement Learning},
  booktitle    = {2020 IEEE Intelligent Vehicles Symposium (IV)},
  pages        = {971--977},
  year         = {2020},
  organization = {IEEE},
  doi          = {10.1109/IV47402.2020.9304721}
}

@article{talia2025incremental,
  author  = {Talia, Sidharth and Salzman, Oren and Srinivasa, Siddhartha},
  title   = {Incremental Generalized Hybrid {A*}},
  journal = {IEEE Robotics and Automation Letters},
  volume  = {11},
  number  = {2},
  pages   = {1586--1593},
  year    = {2025}
}

@article{toupet2026perseverance,
  author  = {Toupet, Olivier and Ono, Masahiro and Sesto, Tyler del and Maimone, Mark and McHenry, Michael},
  journal = {IEEE Transactions on Field Robotics},
  title   = {Enhanced Autonomous Navigation on the {Perseverance} {Mars} Rover},
  year    = {2026},
  volume  = {3},
  pages   = {83--124},
  doi     = {10.1109/TFR.2025.3636366}
}

@inproceedings{richter2024lunar,
  author    = {Richter, J. and Kolvenbach, H. and Valsecchi, G. and Hutter, M.},
  title     = {Multi-Objective Global Path Planning for Lunar Exploration with a Quadruped Robot},
  booktitle = {International Conference on Space Robotics (iSpaRo)},
  pages     = {48--55},
  year      = {2024}
}

@article{patel2024roadrunner,
  author  = {Patel, M. and Frey, J. and Atha, D. and Spieler, P. and Hutter, M. and Khattak, S.},
  title   = {{RoadRunner} {M\&M}: Learning Multi-Range Multi-Resolution Traversability Maps for Autonomous Off-Road Navigation},
  journal = {IEEE Robotics and Automation Letters},
  volume  = {9},
  number  = {12},
  pages   = {11425--11432},
  year    = {2024}
}
\end{document}